\documentclass{article}
\usepackage{iclr2027/iclr2027_conference,times}
\usepackage{amsmath,amssymb,booktabs,microtype,multirow}
\usepackage[final]{graphicx}
\usepackage{hyperref}
\usepackage{url}
\usepackage{xcolor}
\graphicspath{{Figures/}{../Figures/}}

\title{\centering
Grounding Large Language Models\\
in DSGE Simulators for Policy\\
Generation and Forecasting}

\author{%
\parbox{\dimexpr\textwidth-2\tabcolsep\relax}{\centering
Aditya Dubey \hspace{0.65in} Namah Gupta \hspace{0.65in} Vinti Agarwal\\[0.55em]
\normalfont Birla Institute of Technology and Science, Pilani, India}}

\iclrfinalcopy

\begin{document}
\maketitle
\lhead{Preprint}

\begin{abstract}
Large language models can produce economic policy responses that sound
reasonable, but this does not show that their actions are consistent with
economic dynamics. We test this by placing an instruction-tuned language model
inside six Snowdrop-backed dynamic stochastic general equilibrium (DSGE)
simulators. At each
turn, the model observes the economy and a change in economic discourse,
selects a bounded policy action, and receives the next simulated state and an
economic reward. We implement a common Python interface for repeated rollouts,
persistent shocks, state cloning, and rolling-horizon simulation.

This setting creates a long-horizon credit-assignment problem. Policy effects
may appear several quarters after an action is taken. PPO has a learned value
function that can propagate delayed reward to earlier tokens through generalized
advantage estimation. GRPO has no learned value function and instead assigns a
group-relative advantage from complete rollout returns. It therefore cannot
distinguish which earlier turn caused the outcome; if every rollout receives
the same return, the normalized advantage is zero. We use PPO as the primary
method and GRPO as a matched critic-free baseline. The experiments also test
directional semantic signals, reward horizon, trajectory warm starts,
cross-simulator transfer, and historically anchored pandemic and
monetary-policy shocks. The objective is to judge policy actions by their
simulated economic consequences rather than by plausible language alone.

\end{abstract}

\section{Introduction}

An economic policy response can be fluent and well argued while still implying
an unstable or internally inconsistent trajectory. Language models are normally
judged on the response itself; the economic consequences remain outside the
learning loop. We instead use a DSGE simulator as the environment that applies
an action, advances the economy, and returns numerical feedback.

At each turn, the model observes the current state, the policy instruments
available in that simulator, and a change in economic discourse. It returns a
bounded JSON action. The environment combines this action with the scenario
shocks, simulates the next period, and returns the new state and loss. Policy
generation therefore becomes repeated control rather than a one-shot prompt.

This creates two different problems. First, policy transmission is delayed, so
an early action must receive credit for an outcome that appears several turns
later. PPO can use its critic to assign different values within the trajectory.
Vanilla GRPO instead compares complete rollout returns and has no learned value
function for within-trajectory credit. Second, both algorithms are local
optimizers: they cannot learn a useful trajectory that the starting policy
never generates. With an immediate action penalty, this can produce an
\emph{inaction ravine} in which doing little is safer than exploring an
intervention whose benefit arrives later.

We lack a large expert trajectory bank for task-specific SFT. We therefore
start from an instruction-tuned model, separate credit assignment from
exploration, and test a small rule-based trajectory warm start. The semantic
component is also directional: it uses the change between consecutive discourse
windows as a noisy shock signal, not as a causally identified shock.

Our contributions are fivefold:
\begin{enumerate}
\setlength{\itemsep}{1pt}
\setlength{\parskip}{0pt}
\item We define an RL-compatible interface to six Snowdrop models: a quarterly
projection model (QPM), Smets--Wouters (SW), Gal\'i--Smets--Wouters (GSW),
Ireland's New Keynesian model, a U.S. multivariate-filter model, and a
real-business-cycle model. The interface supports state cloning, persistent
shock schedules, rolling forecasts, and irreversible committed transitions.
\item We formulate semantic-shift conditioning with chronological data splits
and named shock heads.
\item We fine-tune the language model with parameter-efficient adapters rather
than training only a shallow action head. This makes cross-simulator transfer a
test of whether policy behavior learned in one structural model carries to
another.
\item We motivate PPO with a learned critic as the primary method for delayed
multi-turn control and retain GRPO as a matched multi-rollout baseline.
\item We separate algorithm, reward, exploration, semantic, and
cross-simulator effects through targeted ablations and external-event
validation.
\end{enumerate}

\begin{figure}[t]
\centering
\includegraphics[width=\linewidth]{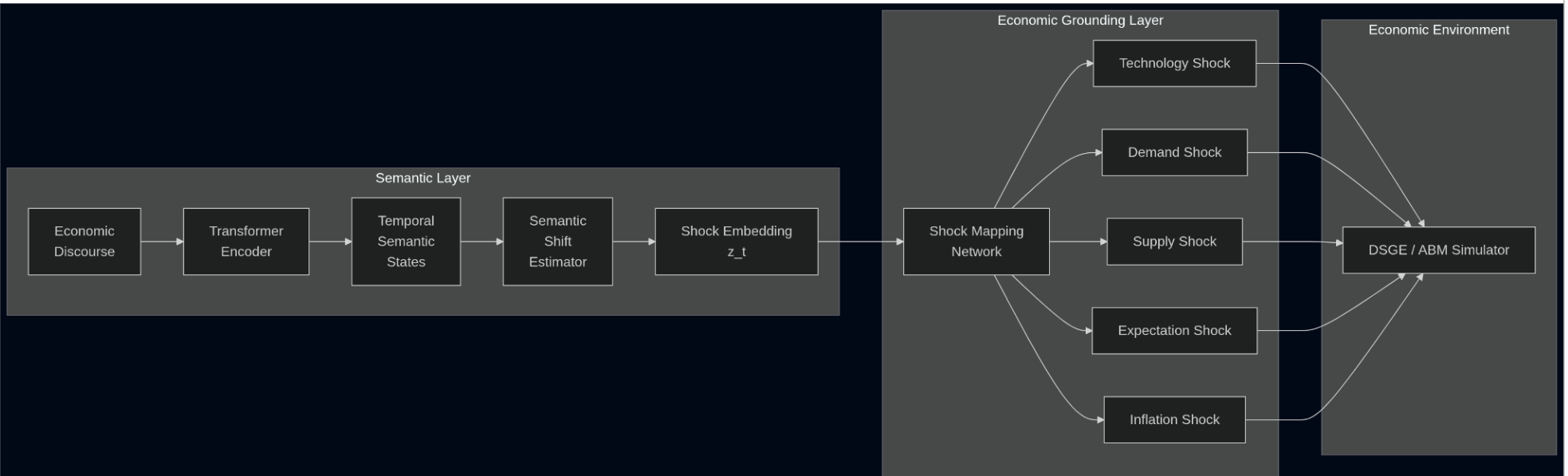}
\caption{Semantic grounding architecture. Economic discourse is encoded into a
temporal semantic state, mapped to named shock hypotheses, and evaluated by a
DSGE simulator rather than by linguistic plausibility alone.}
\label{fig:semantic-architecture}
\end{figure}

\section{Related Work}

\paragraph{Simulator grounding.}
GLAM trains language agents using feedback from an interactive environment, and
bilevel systems let an LLM propose hypotheses whose numerical consequences are
checked by a simulator \citep{carta2023glam,ma2024bilevel}. We use the same idea
for structural macroeconomics. Snowdrop is the Python numerical backend, and
the registered models draw on established DSGE implementations
\citep{Goumilevski2025,smets2007shocks,ireland2004technology}. These simulators
make assumptions and counterfactuals explicit; they are not an oracle for the
real economy.

\paragraph{Multi-turn RL and exploration.}
PPO, RLOO, GRPO, and related methods are commonly used where reward follows one
response. In multi-turn control, the outcome may arrive after several
decisions, breaking this direct action--reward link. PPO combines a critic, GAE,
and clipped updates, whereas GRPO replaces the critic with relative returns
from grouped rollouts
\citep{schulman2016gae,schulman2017ppo,shao2024deepseekmath}. Multi-turn studies
therefore motivate finer credit assignment
\citep{wang2025practitioners,zeng2025multiturn}. This is separate from
exploration. GEP-PG first finds diverse trajectories and then applies policy
gradients; hindsight replay, return redistribution, curiosity, and SAC-GLAM
provide related replay or exploration mechanisms
\citep{colas2018geppg,andrychowicz2017her,arjona2019rudder,pathak2017curiosity,gaven2026sacglamimprovingonlinerl}.

\paragraph{Training stages and semantic signals.}
Standard RLHF combines SFT, reward modeling, and policy optimization
\citep{ouyang2022instructgpt}. Here the DSGE loss replaces a learned preference
reward. SFT can expand the set of correct reasoning paths, while RL concentrates
probability on paths the model already produces
\citep{matsutani2025squeezes}. This motivates the trajectory warm-start
ablation. Generic embeddings can miss small changes in financial narratives,
and discourse may itself respond to market conditions
\citep{liu2024semantic,drinkall2025stories}. We therefore model semantic change
as an observation and leave causal identification to the structural model and
evaluation design.

\section{Grounded Policy Environment}
\label{sec:environment}

At turn $t$, the hidden economic state is $x_t$, the observed state is $o_t$,
and tokens $a_{t,1:L_t}$ encode policy action $u_t$. Simulator $m$ advances as
\begin{equation}
  x_{t+1}=F_m(x_t,u_t,\varepsilon_t;\theta_m), \qquad
  o_{t+1}=H_m(x_{t+1}),
\end{equation}
where $\theta_m$ is its calibration and $\varepsilon_t$ is the scenario shock.
The LLM does not generate equations; it chooses from policy levers exported by
the selected simulator.

\subsection{Multi-rollout DSGE interface}

Snowdrop supplies the numerical solvers and structural model implementation
\citep{Goumilevski2025}; the RL interface is our implementation. Our wrapper
turns these solvers into a stateful environment for repeated, matched
language-agent rollouts. The wrapper adds four capabilities needed here: (i) shock abstraction,
which accepts scalar, full-path, or sparse time--value schedules; (ii) named
parameter updates for controlled regime experiments; (iii) deep state cloning,
which copies the calibration, committed state, active shocks, and loss state;
and (iv) an RL step that validates an action, advances the simulator, and returns
the next observation, loss, reward, termination flag, and validity flag.
Rolling forecasts and irreversible one-period commits preserve the DSGE
temporal structure while allowing many counterfactual trajectories to begin
from the same economy.

For GRPO, the environment is cloned $G$ times and one complete trajectory is
sampled from each clone before group-relative advantages are computed. PPO
retains the trajectory and generation-time token probabilities while a separate
critic supplies token values. Both algorithms therefore receive the same
initial state and exogenous shock path.

\subsection{Validated actions and irreversible transitions}

An action is a JSON object with a lever, signed magnitude, and optional short
rationale. Model-specific bounds are enforced before simulation. Invalid JSON,
unknown levers, and out-of-range magnitudes incur a penalty rather than being
silently repaired. A policy shock persists for a fixed duration with geometric
decay. The environment computes a rolling forecast, commits only the first
transition, and appends it to immutable realized history. Consequently, a later
turn cannot rewrite earlier outcomes.

Table~\ref{tab:models} shows the current registry. Targets and levers are
model-specific; this prevents an action valid in one model from being
mistakenly interpreted in another.

\begin{figure}[t]
\centering
\includegraphics[width=\linewidth]{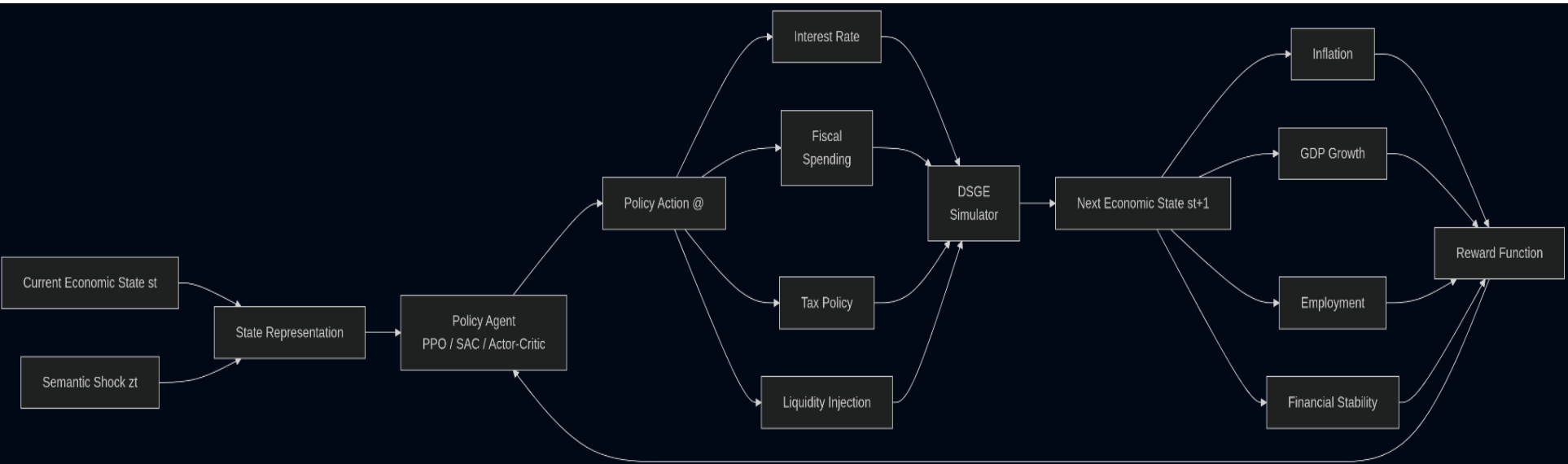}
\caption{Closed-loop policy grounding. A semantic shift conditions the policy
agent, a validated action modifies the DSGE shock path, and multivariate
simulator outputs determine the reward and next observation.}
\label{fig:policy-loop}
\end{figure}

\begin{table}[t]
\caption{Structural models exposed through the common environment.}
\label{tab:models}
\centering
\small
\begin{tabular}{@{}p{0.35in}p{1.05in}p{1.65in}p{1.55in}@{}}
\toprule
Key & Structural model & Principal targets & Policy lever(s) \\
\midrule
QPM & Quarterly projection & inflation, output gap & monetary, fiscal \\
SW & Smets--Wouters & inflation, output & monetary, fiscal \\
GSW & Gal\'i--SW pandemic & inflation, output, unemployment & monetary, fiscal \\
IRE & Ireland (2004) & inflation, output gap & monetary \\
MVF & U.S. multivariate filter & inflation, output, unemployment & demand stabilizer \\
RBC & Real business cycle & output, consumption & output stabilizer \\
\bottomrule
\end{tabular}
\end{table}

\subsection{Reward and policy objective}

For target variable $j$ with desired value $y_j^\star$ and weight $w_j$, the
planning-mode loss over a simulated horizon $H$ is
\begin{equation}
 \mathcal{L}_t=\sum_{h=1}^{H}\sum_j
 w_j\left(y_{t+h,j}-y_j^\star\right)^2,
\end{equation}
while control mode applies the same weighted loss to the newly committed state.
The implemented normalized improvement is $\Delta_t=(\mathcal{L}_{t-1}-
\mathcal{L}_t)/\max(|\mathcal{L}_0|,1)$. The bounded reward is
\begin{equation}
 r_t=c\tanh(\Delta_t)-\rho\lVert u_t\rVert_2^2
 -p_{\mathrm{invalid}}-p_{\mathrm{semantic}},
 \label{eq:reward}
\end{equation}
where the last term is an optional terminal penalty under high measured
semantic volatility. Because sequential improvement can differ from the same-state causal effect
of an action, the confirmatory evaluation will replay matched no-action
trajectories from identical initial states. It will report both reward and the
underlying economic loss so reward shaping cannot hide poor outcomes.

The quadratic objective is a baseline, not a claim that reward design is
settled. Policy costs appear immediately, while stabilization may take several
quarters. This can make inaction locally attractive before the policy discovers
a successful intervention, a reward geometry similar to the deceptive gradient
studied in Continuous Mountain Car \citep{colas2018geppg}. We therefore vary the
forecast horizon, tolerance bands around targets, asymmetric inflation/output
penalties \citep{cukierman2008nonlinear}, action costs, and the semantic-volatility
penalty. Economic loss is always reported separately from shaped reward.

\section{Semantic Shifts as Economic Observations}
\label{sec:semantic}

Let $D_{t-1}$ and $D_t$ be consecutive windows of economic discourse. A
transformer encoder $f_\phi$ produces document states $z_{t-1}$ and $z_t$:
\begin{equation}
 s_t=\left[z_t;\;\frac{z_t-z_{t-1}}{\lVert z_t-z_{t-1}\rVert_2};\;
 \cos(z_t,z_{t-1})\right].
\end{equation}
A projection $g_\psi(s_t)$ feeds named shock heads. Training uses shock
regression, semantic-volatility prediction, and temporal consistency. At policy time, this representation or a precomputed semantic embedding is
passed through separate actor and critic bridges.

The direction $z_t-z_{t-1}$ is the main signal: a static embedding identifies
what a document discusses, while the difference records how the discussion
changed. We interpret it only as a noisy shock observation. Chronological
splits prevent future discourse from entering past policy states.

\begin{figure}[t]
\centering
\includegraphics[width=0.72\linewidth]{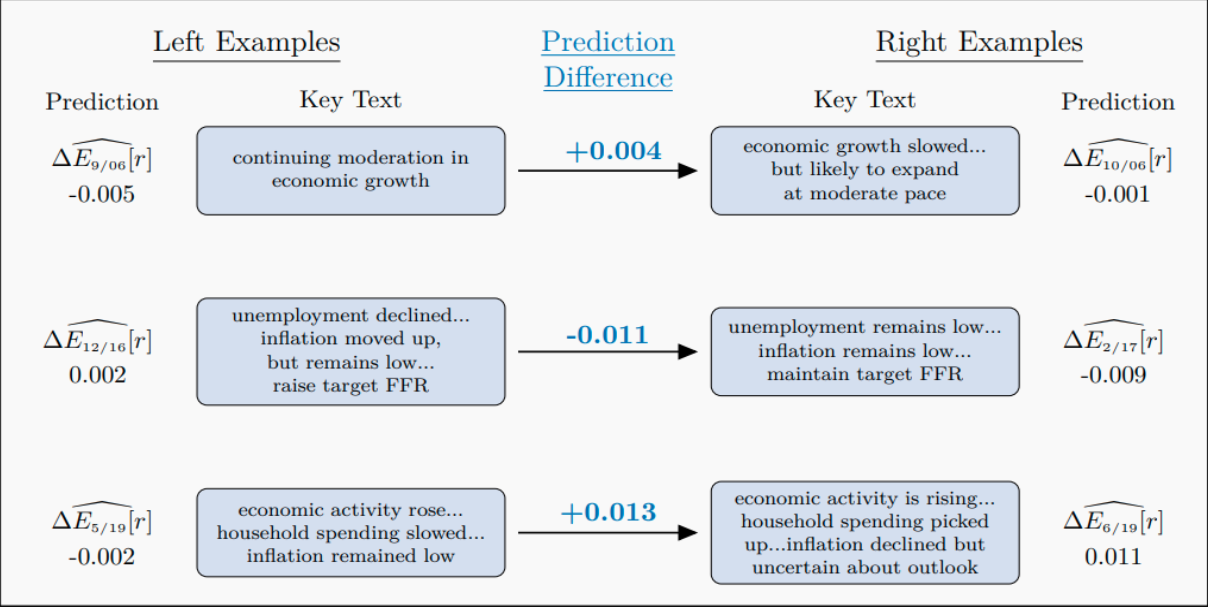}
\caption{Paired financial narratives with small but decision-relevant semantic
differences. Adapted from the Financial-STS examples of
\citet{liu2024semantic}; the examples motivate supervision beyond static topic
similarity.}
\label{fig:semantic-pairs}
\end{figure}

\section{Credit Assignment Across Turns}
\label{sec:credit}

The central question is how a delayed reward reaches the decisions that caused
it. Suppose an action is a token sequence that the environment executes only at
the command boundary. Following \citet{wang2025practitioners}, the turn reward
is assigned only to the end-of-sequence token:
\begin{equation}
r_t^i=\begin{cases}
r_t, & a_t^i=\texttt{<eos>},\\
0, & \text{otherwise}.
\end{cases}
\end{equation}
For token $i$ in turn $t$, PPO then uses
\begin{align}
 \delta_t^i &= r_t^i+\gamma V(h_t^{i+1})-V(h_t^i),\\
 \widehat A_t^i&=\sum_{\ell=0}^{L-i}(\gamma\lambda)^\ell
 \delta_t^{i+\ell}.
 \label{eq:gae}
\end{align}
This is why PPO is useful here: although only the command-ending token receives
the reward, the critic and GAE can propagate non-zero advantages to preceding
tokens. With probability ratio $q_t^i(\theta)$, the actor maximizes
\begin{equation}
 \mathcal{J}_{\mathrm{PPO}}=\mathbb{E}\left[\sum_{t,i}
 \min\left(q_t^i\widehat A_t^i,
 \mathrm{clip}(q_t^i,1-\epsilon,1+\epsilon)\widehat A_t^i\right)\right].
\end{equation}

Vanilla trajectory-level GRPO samples $G$ rollouts for a prompt and uses
\begin{equation}
 \widehat A_g=\frac{R_g-\overline R}{\mathrm{sd}(R_{1:G})+\epsilon}.
 \label{eq:grpo}
\end{equation}
Every selected token in rollout $g$ receives this same group-relative signal.
GRPO can therefore reinforce a successful rollout, but without a learned value
function it cannot distinguish which earlier turn caused the final outcome. If
$R_1=\cdots=R_G$, Eq.~\ref{eq:grpo} becomes zero and the whole group supplies no
task gradient. When returns differ, stochastic variation can still create a
noisy relative ranking. Dense turn rewards, clearly separated terminal returns,
or turn-level variants such as MT-GRPO can restore a more informative signal.

\paragraph{Evidence motivating the comparison.}
Two observations determine our primary comparison. First, PPO's critic can
estimate which parts of a long decision sequence require updating. The
multi-turn formulation of \citet{wang2025practitioners} makes the mechanism
explicit: value bootstrapping gives tokens preceding a command-ending reward
non-zero advantages. On their Qwen-1.5B TextWorld experiments, PPO reaches
$59\%$ success versus GRPO's $2\%$ on w4-o6-q8, and $88\%$ versus $18\%$ on
w2-o3-q4. At 7B, PPO still leads GRPO by $72\%$ to $36\%$ on the harder task,
while the gap narrows to $98\%$ versus $79\%$ on the simpler task. These are
results from the cited TextWorld study, not from our DSGE environment.

Second, GRPO's main learning signal depends on the return distribution within a
group of stochastic rollouts. Equal returns give zero advantage, while incidental
variation can make the relative ranking noisy; group-relative estimation can
also weight easy and hard prompts unevenly \citep{yang2026biased}. We therefore
hypothesize that vanilla GRPO becomes less reliable as DSGE transmission lags
increase. This is not a universal claim against GRPO. It remains attractive when
a terminal verifier sharply separates rollouts, as in the cited SWE-Gym setting,
and turn-level variants can restore finer credit
\citep{wang2025practitioners,zeng2025multiturn}. Our matched cloned-state design
tests whether this ordering actually transfers to economic control.

\paragraph{Credit assignment is not exploration.}
A critic can propagate reward along an observed trajectory, but it cannot
assign credit to a useful trajectory that the policy never generates. This is
especially relevant when training starts from an instruction model without a
large domain-specific SFT corpus. We therefore interpret PPO--GRPO as a
credit-assignment comparison and separately test exploration initialization.
A rule-based controller with randomized perturbations builds a small library of
valid, non-trivial trajectories; warm-starting from this library follows the
explore-then-optimize logic of GEP-PG \citep{colas2018geppg}. Replay-based
hindsight learning and return redistribution are reserved as extensions because
they alter the on-policy objective rather than merely its initialization.

\section{Datasets and Scenario Construction}
\label{sec:data}

Controlled scenarios from the six registered DSGE models provide training
and mechanistic tests under fixed initial states and shock paths. They cover
recession, stagflation, cost-push, productivity, lockdown, labor-supply, and
potential-output disturbances. External data are reserved for evaluation.

The McKibbin--Fernando release provides six COVID-19 scenarios, their source
shocks, and G-Cubed outcomes
\citep{mckibbin2020scenarios,repec:eee:ecmode:v:129:y:2023:i:c:s0264999323003632}.
Because G-Cubed is not our GSW model, a documented adapter maps its productivity,
demand, risk, fiscal, morbidity, and mortality quantities to quarterly GSW shock
paths; Scenario 6 is marked as approximate. The planned comparison reports output/inflation RMSE and correlation
separately from policy-loss improvement against a cloned no-action GSW
trajectory.

The ECB evaluation uses meeting-level and monthly Jaroci\'nski--Karadi shocks
derived from EA-MPD event windows
\citep{altavilla2019measuring,jarocinski2020deconstructing}.
Only the identified monetary component enters the policy-shock channel; the
central-bank-information component remains an observation. The planned metrics are action direction, response timing, and stabilization
loss, not agreement with the historical ECB decision.

\begin{table}[t]
\caption{Data sources and their non-overlapping roles in the evaluation.}
\label{tab:data-sources}
\centering
\small
\resizebox{\linewidth}{!}{%
\begin{tabular}{llll}
\toprule
Source & Contents & Role & Primary metrics \\
\midrule
DSGE registry & controlled shocks, states, discourse & training and ablations & return, loss, validity \\
COVID-19 release & scenario shocks and G-Cubed outcomes & structural stress test & RMSE, correlation, loss improvement \\
ECB/EA-MPD & meeting and monthly policy surprises & historical event validation & sign, timing, stabilization loss \\
\bottomrule
\end{tabular}}
\end{table}

Controlled scenarios are split by shock family and simulator, semantic data are
split chronologically, and both external sources remain locked until model and
reward selection are complete.

\section{Experimental Protocol}
\label{sec:experiments}

\subsection{Research questions and hypotheses}

We evaluate six questions. \textbf{RQ1:} Does grounded RL reduce economic
loss relative to the base LLM, random valid actions, no action, and a simple
rule-based controller? \textbf{RQ2:} Does PPO outperform trajectory-level GRPO
as horizon and scenario difficulty increase? \textbf{RQ3:} Does a temporal
semantic signal improve shock recognition and policy choice over static text or
state-only control? \textbf{RQ4:} Which reward geometry best represents delayed
stabilization rather than immediate target chasing? \textbf{RQ5:} Does a small
library of exploratory rule-based trajectories improve learning from the
instruction model? \textbf{RQ6:} Do policies learned in one simulator retain
value under held-out models and real-event shock paths?

The hypotheses are: (H1) grounded policies outperform linguistic and ungrounded
baselines; (H2) PPO's margin over vanilla GRPO grows with delayed feedback and
its critic propagates terminal reward to earlier tokens; (H3) directional
semantic conditioning improves shock classification and downstream economic
loss; (H4) rolling-horizon and tolerance-band rewards produce more informative
policy rankings than one-step exact-target loss; (H5) exploration warm-starting
improves both action diversity and final economic loss; and (H6) multi-model
training improves held-out-simulator robustness but cannot remove model
misspecification.

\subsection{Training and baselines}

The policy is Qwen2.5-1.5B-Instruct \citep{qwen2025technical}, quantized to four
bits and adapted with LoRA \citep{hu2022lora}. PPO uses a separate token-value
model, $\gamma=0.99$, $\lambda=0.95$, policy clip $0.2$, and four PPO epochs.
GRPO samples four trajectories per prompt, each from a clone of the same
economic state, normalizes total returns within the
group, applies a $0.2$ ratio clip, and uses a reference-policy KL penalty. Each
episode has four decision turns and an eight-period forecast horizon.
Confirmatory runs will retain seeds, scenario order, states, actions, per-turn
rewards, economic losses, parse failures, and checkpoints.

Baselines are: (i) no intervention; (ii) uniformly sampled valid action; (iii)
the untuned instruction model; (iv) a model-specific linear stabilization rule;
(v) GRPO; and (vi) PPO. We additionally report state-only PPO, static-text PPO,
temporal-semantic PPO, and an oracle-shock PPO upper bound. RLOO and MT-GRPO are
extensions, separated from the core table unless their implementations pass the
same rollout-accounting tests.

We update LoRA parameters throughout the compact language model rather than
training only an action-classification head. This choice makes the transfer test
meaningful: success requires the language policy to reuse its learned mapping
from economic observations and discourse to actions under a different set of
structural equations. It does not imply that the model recovers those equations
or identifies causal shocks from text.

The pipeline therefore differs from full preference-based RLHF: general
instruction tuning supplies the initialization, the DSGE loss replaces a
learned reward model, and LoRA PPO or GRPO performs the task update. For the
proposed warm-start condition, a dataset-to-trajectory translator will record
each source shock, converted path, state, legal action, next state,
and horizon outcome. Randomized rule-based policies will populate a small
trajectory library; direct-RL runs will omit it. This tests whether better
initial trajectory coverage is required in addition to simulator feedback.

\subsection{Ablations, metrics, and controls}

The predeclared ablations isolate credit assignment (PPO versus GRPO), horizon
(one step versus 4/8/12 quarters), reward geometry (quadratic versus
tolerance-band or asymmetric), exploration (direct RL versus rule-based
warm-start), semantics (none, static, directional, or oracle), actor/critic
conditioning, and single- versus multi-model training. This separation prevents
an aggregate score from attributing an exploration or reward-design effect to
the optimizer.

Primary outcomes are discounted return and target-weighted economic loss.
Secondary outcomes are policy energy, worst-case loss, action validity, return
variance, zero-advantage group frequency, shock-label macro-F1, and calibration
of predicted semantic volatility. Each comparison uses identical shock paths
and initial states. The confirmatory tables will report means, standard errors,
and paired bootstrap confidence intervals across seeds and scenarios. Model selection will
use validation scenarios only; external events will be evaluated once after
selection.

\section{Empirical Results}
\label{sec:initial-results}

We organize the evidence in the same order as the causal questions in the
protocol: simulator validity, algorithmic credit assignment, reward geometry,
semantic conditioning, and transfer. Previously recorded simulator checks and the
calibration diagnostic retain their measured values. Tables for the
confirmatory policy-learning runs use \emph{TBD} cells until their logged
artifacts are inserted; these cells are not numerical claims.

\subsection{Simulator validation}

We first removed the language model and tested the simulator interface directly
on the linear QPM and nonlinear SW models. A ten-case suite covered isolated
shocks, combined shocks, persistence, and mean reversion. For a unit nominal
interest-rate shock in the QPM, the initial simulated values were
$RS=16.0100$, $RR=8.4612$, and annualized inflation $DLA\_CPI=7.2507$; one
quarter later they moved to $13.2724$, $11.0044$, and $5.5257$. In the nonlinear
SW multiple-shock test, output, inflation, and the policy rate moved from the
steady state to $(-0.035428,-0.007073,0.020122)$ after one quarter. These
responses established that shocks reached the intended equations and that the
wrapper returned non-trivial transitions.

\begin{table}[t]
\caption{Representative outputs from the ten-case simulator verification
suite. Values are reproduced from the recorded QPM and SW verification output.}
\label{tab:simulator-validation}
\centering
\small
\begin{tabular}{@{}llrrr@{}}
\toprule
Model/test & Period & \multicolumn{3}{c}{Simulated variables} \\
\cmidrule(l){3-5}
QPM, $SHK\_RS=1.0$ & & $RS$ & $RR$ & $DLA\_CPI$ \\
 & Initial & 16.010000 & 8.461200 & 7.250700 \\
 & Next quarter & 13.272388 & 11.004444 & 5.525716 \\
\midrule
SW, multiple shocks & & $y$ & $pinf$ & $r$ \\
 & Initial & 0.000000 & 0.000000 & 0.000000 \\
 & Next quarter & $-0.035428$ & $-0.007073$ & 0.020122 \\
\bottomrule
\end{tabular}
\end{table}

\subsection{Matched PPO--GRPO comparison}

The primary comparison holds the model, scenarios, initial states, action
schema, number of decision turns, and total environment interactions fixed.
Only the advantage estimator changes. The directional hypothesis is that PPO's
advantage grows as the forecast and reward delay increase: its critic can assign
different values within a trajectory, whereas vanilla GRPO relies on return
differences among complete cloned rollouts. We test the mechanism directly by
reporting the fraction of action tokens with non-zero advantage for PPO and the
fraction of zero-variance rollout groups for GRPO.

\begin{table}[t]
\caption{Confirmatory PPO--GRPO comparison. Values will be inserted from saved
rerun artifacts; lower economic loss and higher return are better.}
\label{tab:ppo-grpo-results}
\centering
\small
\resizebox{\linewidth}{!}{%
\begin{tabular}{@{}llrrrr@{}}
\toprule
Horizon & Method & Return & Econ. loss & Valid action & Credit diagnostic \\
\midrule
Short & Base LLM & TBD & TBD & TBD & -- \\
Short & GRPO & TBD & TBD & TBD & TBD \\
Short & PPO & TBD & TBD & TBD & TBD \\
Long & Base LLM & TBD & TBD & TBD & -- \\
Long & GRPO & TBD & TBD & TBD & TBD \\
Long & PPO & TBD & TBD & TBD & TBD \\
\bottomrule
\end{tabular}}
\end{table}

Our interpretation is conditional on these diagnostics. A PPO improvement
accompanied by broader non-zero token advantages supports the temporal-credit
explanation. A GRPO deficit accompanied by frequent equal-return groups supports
the group-signal explanation. A difference without either diagnostic would not
establish the proposed mechanism.

\subsection{Semantic, reward, exploration, and transfer ablations}

The semantic ablation progresses from economic state only, to a static text
embedding, to the directional shift $z_t-z_{t-1}$, and finally to oracle shock
labels. Static text tests whether topic information is sufficient; the
directional condition tests the paper's central claim that the movement of
discourse is the informative quantity; the oracle condition bounds loss caused
by semantic measurement error. Conditioning only the actor is also compared
with conditioning both actor and critic.

The reward ablation crosses short and long forecast horizons with quadratic,
tolerance-band, and direction-dependent losses. The exploration ablation
compares direct RL from Qwen2.5-1.5B-Instruct with a warm-start library generated
by randomized rule-based policies. The latter does not replace RL: it tests
whether first discovering economically useful trajectories improves subsequent
policy-gradient optimization. Finally, single-simulator training is compared
with multi-model training on held-out structural models.

\begin{table}[t]
\caption{Directional summary of the confirmatory ablations. Numeric estimates
and confidence intervals remain to be populated from the rerun.}
\label{tab:ablation-results}
\centering
\small
\begin{tabular}{@{}p{1.2in}p{2.15in}p{1.65in}@{}}
\toprule
Question & Primary contrast & Hypothesized direction \\
\midrule
Semantic shift & directional vs. static text & higher shock F1; lower loss \\
Reward horizon & rolling vs. one-step & fewer policy-ranking reversals \\
Reward geometry & tolerance band vs. exact target & greater target-zone occupancy \\
Exploration & warm-start vs. direct RL & more useful actions; lower final loss \\
Transfer & multi-model vs. single-model & lower held-out-model loss \\
\bottomrule
\end{tabular}
\end{table}

\subsection{Reward-horizon calibration diagnostic}

We isolate reward design from language-model learning by replacing the policy
with manually selected actions. Table~\ref{tab:grid-search} reports the measured
base and action losses under the original short-horizon objective. Negative
$\Delta$ means that the action increased that particular objective; it does not
by itself establish that the intervention is economically invalid over a longer
horizon. The sensitivity of the sign to action magnitude and target weights is
the reason horizon and reward geometry are varied in the confirmatory protocol.

\begin{table}[t]
\caption{Calibration grid search from the initial DSGE experiments.}
\label{tab:grid-search}
\centering
\small
\resizebox{\linewidth}{!}{%
\begin{tabular}{lclrrrr}
\toprule
Scenario / action & Magnitude & Weighting & Base loss & Action loss & $\Delta$ & Direction \\
\midrule
Recession / cut & 0.05 & Standard & 216 & 219 & $-3.0$ & worse \\
Recession / cut & 0.50 & Standard & 216 & 248 & $-32.0$ & worse \\
Stagflation / hike & 0.05 & Hawk & 334 & 333 & $+1.6$ & better \\
Stagflation / hike & 0.50 & Standard & 369 & 378 & $-9.6$ & worse \\
Cost-push / hike & 0.25 & Standard & 2664 & 2996 & $-331.8$ & worse \\
\bottomrule
\end{tabular}}
\end{table}

We call this a \emph{reward-horizon inversion}: an intervention with a plausible
long-run stabilizing interpretation can receive a worse short-horizon score.
Under stagflation, even the sign of the comparison depends on the relative
inflation and output weights. This is a property of the evaluated objective,
not evidence for or against a particular RL optimizer. It motivates rolling
forecast loss, matched no-action counterfactuals, persistent actions, and the
predeclared horizon and reward-geometry ablations.

\section{Extensions and Research Directions}
\label{sec:extensions}

The main experiments keep PPO, GRPO, the action schema, and environment
interactions comparable. The next extensions address the two remaining
problems: discovering useful trajectories and locating delayed economic credit.

\paragraph{Exploration before optimization.}
The rule-based warm start is a small goal-exploration stage. Randomized
Taylor-style controllers search for trajectories that improve on matched
no-action paths; policy-gradient training then concentrates probability on the
useful region \citep{colas2018geppg}. The same archive could support a compact
task-specific SFT stage. Curiosity may widen coverage
\citep{pathak2017curiosity}, but novelty must be restricted to economically
relevant state changes.

\paragraph{Replay and off-policy learning.}
A DSGE transition can be reproduced from the simulator version, calibration,
initial state, action, and shock path. This makes rare successful trajectories
reusable. An off-policy extension based on SAC-GLAM would use an
action-conditioned critic $Q(o,a)$, entropy regularization, and hindsight goals
such as target bands reached by the trajectory
\citep{gaven2026sacglamimprovingonlinerl,andrychowicz2017her}. Replay records must retain model,
calibration, and reward hashes so that stale returns are not reused under a
different environment.

\paragraph{Delayed reward and structural intent.}
RUDDER-style decomposition could move long-horizon return toward the decisions
that produced it \citep{arjona2019rudder}. It must be compared with simpler
rolling rewards, tolerance bands, and asymmetric penalties using held-out
economic loss. A later interface could also let the model state a policy intent,
regime, direction, intensity, and duration. A deterministic compiler would map
that intent to legal simulator shocks, while trace-state observations expose
recent actions, active shock schedules, forecasts, and violations
\citep{ma2024bilevel}. The simulator would retain numerical authority.

\section{Limitations, Safety, and Reproducibility}

A DSGE-grounded answer is conditional on equations, calibration, measurement,
and the admissible action space. It is not a policy recommendation for a real
government or central bank. Model uncertainty is therefore a first-class
output: we report results per simulator and worst-case performance rather than
collapsing disagreement into a single authoritative score. External shock
mappings are approximate and will be released with units, signs, and scaling
choices. Text corpora can reflect media attention and institutional bias;
chronological evaluation limits leakage but not representation error.

For full-scale training, the simulator layer must pass replay invariance,
lag-state continuity, no-action counterfactual equality, shock-sign, and unit
tests. The semantic policy must also pass a behavior-policy log-probability
replay test before PPO updates are interpreted. These checks matter because a
syntactically successful simulation can still implement the wrong transition.
On the available RTX 2080 Ti (11GB), the 1.5B actor fits with four-bit LoRA;
PPO's separate critic is the tighter memory constraint, so rollouts and updates
are serialized and semantic embeddings are precomputed. We use FP16 on this
hardware and reserve larger-model scaling for separate compute.

Code and experimental configurations are available at
\url{https://github.com/vasudeywos/Grounding_LLM_in_DSGE_Simulators_for_Policy_Generation_and_Forecasting.git}.

Without task-specific SFT, on-policy RL can refine only behaviors
explored by the instruction model; the warm-start ablation measures,
but does not eliminate, this limitation.

\section{Conclusion}

We treat policy generation as repeated interaction with an explicit economic
model. The contribution is a common control interface, not a new DSGE model:
Snowdrop remains the numerical backend. The language model chooses a bounded
action, the simulator determines its conditional consequences, and the next
decision starts from the committed economic state. A temporal encoder adds the
direction of discourse change as a noisy shock observation.

The experiment design separates the main claims. PPO versus GRPO tests
multi-turn credit assignment; reward and horizon ablations test the economic
objective; trajectory warm starts test exploration; and semantic and
cross-simulator ablations test whether the policy uses directional language and
transfers across structural assumptions. Pandemic and ECB shock paths provide
historically anchored evaluation. The final criterion is not whether a policy
sounds convincing, but whether it improves clearly reported economic outcomes
under explicit model assumptions.

\bibliography{references}
\bibliographystyle{iclr2027/iclr2027_conference}

\appendix
\section{Reproducibility Checklist}

An experiment is complete only when the following artifacts are retained: the git
commit, Python and CUDA environment, simulator YAML hashes, dataset versions,
seeds, prompts, adapter and critic checkpoints, per-token behavior log
probabilities, per-turn state/action/reward records, counterfactual losses, and
the script that regenerates every table. Confirmatory aggregates will be populated through the artifact-generation
script rather than manual editing.

\section{Action Schema}

The current control action is constrained to the following logical schema:
\begin{verbatim}
{"lever": <model-specific lever>,
 "magnitude": <bounded signed float>}
\end{verbatim}
Only these parsed fields affect the transition. Confirmatory runs will retain
the raw generation for auditability, but it receives no direct reward. This
prevents eloquence from changing the simulated outcome.

\section{Result Aggregation Protocol}

For each algorithm and seed, evaluation will begin from the same serialized
state and shock path. Raw episode records will be exported before aggregate
statistics are computed in a fresh process. A result cell will be populated only
when its records, configuration, and checkpoint are present. Failed runs will
remain in validity and failure-rate metrics.

\end{document}